\documentclass{article}
\usepackage{spconf,amsmath,graphicx}
\usepackage{multirow}
\usepackage{ctable}
\usepackage{caption}
\usepackage{subcaption}

\title{IOD-CNN: Integrating Object Detection Networks \\ for Event Recognition}

\name{\textsuperscript{$\star$}Sungmin Eum\textsuperscript{$\dagger\ddagger$}, \textsuperscript{$\star$}Hyungtae Lee\textsuperscript{$\dagger\mathsection$}, Heesung Kwon\textsuperscript{$\dagger$}, David Doermann\textsuperscript{$\ddagger$} \thanks{\textsuperscript{$\star$}These authors contributed equally to this work}}
\address{
         \textsuperscript{$\dagger$}U.S. Army Research Laboratory, Adelphi, MD, USA\\
         \textsuperscript{$\ddagger$}Institute of Advanced Computer Studies, University of Maryland, College Park, MD, USA\\
         \textsuperscript{$\mathsection$}Booz Allen Hamilton Inc., McLean, VA, USA
         }
\begin{document}
%
\maketitle
\begin{abstract}
Many previous methods have showed the importance of considering semantically relevant objects for performing event recognition, yet none of the methods have exploited the power of deep convolutional neural networks to directly integrate relevant object information into a unified network. We present a novel unified deep CNN architecture which integrates architecturally different, yet semantically-related object detection networks to enhance the performance of the event recognition task. Our architecture allows the sharing of the convolutional layers and a fully connected layer which effectively integrates event recognition, rigid object detection and non-rigid object detection. 

\end{abstract}
\section{Introduction}
\label{sec:Introduction}
To better perform event or action recognition, recently introduced approaches have exploited the importance of  considering semantically relevant and distinctive objects. For example, Althoff et al. \cite{TAlthoffACMMM2012} showed that the statistics derived from object detection results can better represent events. Joel et al. \cite{JLevisArxiv2016} claimed that event recognition performance can be enhanced by incorporating semantically related keywords which represent the salient objects. Jain et al. \cite{MJainCVPR2015} showed that objects do matter for actions by encoding object categories that benefit the action recognition as well as object localization.

Recently, Wang et al. \cite{LWangCVPRW2015} presented an approach which uses two separate deep convolutional neural networks (CNNs), an object CNN and a scene CNN. They used a simple late fusion to combine the fully connected (FC) layer outputs from the networks and applied a support vector machine (SVM) for classification. An enhanced network was introduced in \cite{LWangICCVW2015} by incorporating the local features (TDD: Transformed Deep-convolutional Descriptor) because the features from the FC layers were found to be weak in capturing the local information in the images. Both approaches use separate networks which are integrated with a late fusion.

In our approach, we exploit the power of deep convolutional neural networks (CNNs) in combining different networks (for different tasks) together in an end-to-end multi-task learning scheme. Learning a unified network allows better harvesting of the semantically relevant object information to boost event recognition. We incorporate event recognition as a primary task and relevant object detections as secondary tasks. This approach is motivated by previously methods \cite{SKimICIP2016,ZZhangECCV2014,SLiIJCV2015,ZKangICML2011,RCaruanaML1997} which have demonstrated that a task can be better learned assisted by appropriate secondary tasks. 

There are several technical challenges in constructing a unified deep network which integrates image classification (event recognition in our case) and object detection which are architecturally different in nature. First, the image classification system must pass an input image through the sequential layers of a network and generate class probability scores as an output \cite{AkrizhevskyNIPS2012,KSimonyanICLR2015,CSzegedyCVPR2015,KHeCVPR2016}. Second, object detection must generate local candidate object region of interests (RoIs) which are evaluated to compute their scores. We inherit a widely used object detection approach called the Fast R-CNN \cite{RGirshickICCV2015} for this. This object detection approach uses RoI generation and RoI pooling steps which are the two primary differences when compared to the aforementioned image classification.

To integrate these architectures, we devised a unified CNN framework which enables the sharing of the convolutional layers, one FC layer and one RoI pooling layer between image classification and object detection. As the CNN is integrated by object detection modules, we call it the Integrated Object Detection (IOD)-CNN. The fact that the image classification also uses the RoI pooling layer (which is different from typical image classification) not only makes the network differ in appearance, but also adds a beneficial functionality. With the help of the shared RoI pooling layer, it is no longer necessary to resize the input images to a fixed size. This allows the use of high-resolution images as input, providing room for classification performance enhancement.

For image classification, the input to the RoI pooling (i.e., RoI), is set to be the entire region of the input image. For object detection, object proposals generated by the selective search (rigid objects) or by the multi-scale sliding window search (non-rigid objects) are used as inputs to the RoI pooling.

Our contributions can be summarized as:
\begin{enumerate}
    \item The introduction of a novel unified deep CNN architecture which integrates architecturally different, yet semantically-related networks for different secondary tasks to enhance the performance of a primary task
    \item A demonstration of the effectiveness of the novel approach by showing that the performance of event recognition (primary task) can be boosted by incorporating rigid and non-rigid object detection.
    \item The fact that our architecture can be further enhanced by appending a late fusion, indicating that early-sharing of the layers is complementary to the late fusion.
\end{enumerate}

\section{Our Approach}
\label{sec:OurApproach}

\subsection{IOD-CNN: Integrated Object Detection CNN}
\label{ssec:IODCNN}
In this section, we elaborate on three tasks along with modifications we made to architectures which implement them. We then explain how these different architectures are integrated into a unified network.\\

\begin{figure}[htb]
\begin{subfigure}[htb]{\linewidth}
\includegraphics[trim=7mm 15mm 7mm 5mm,width=\linewidth]{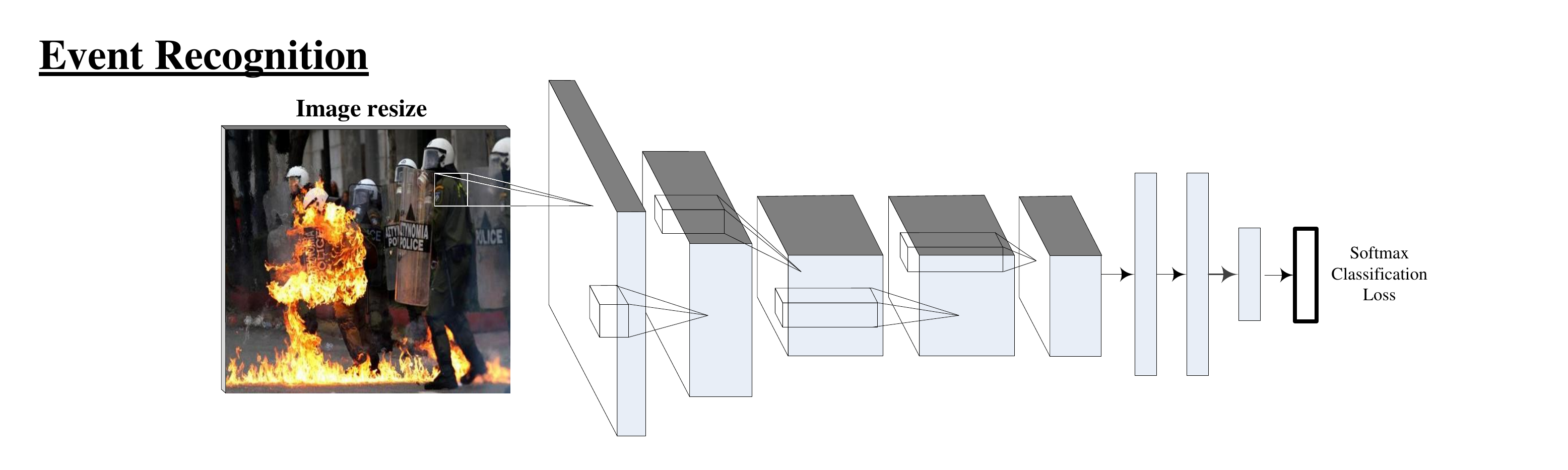}
\subcaption{}
\label{fig:event_recognition}
\end{subfigure}
\begin{subfigure}[htb]{\linewidth}
\includegraphics[trim=7mm 15mm 7mm 5mm,width=\linewidth]{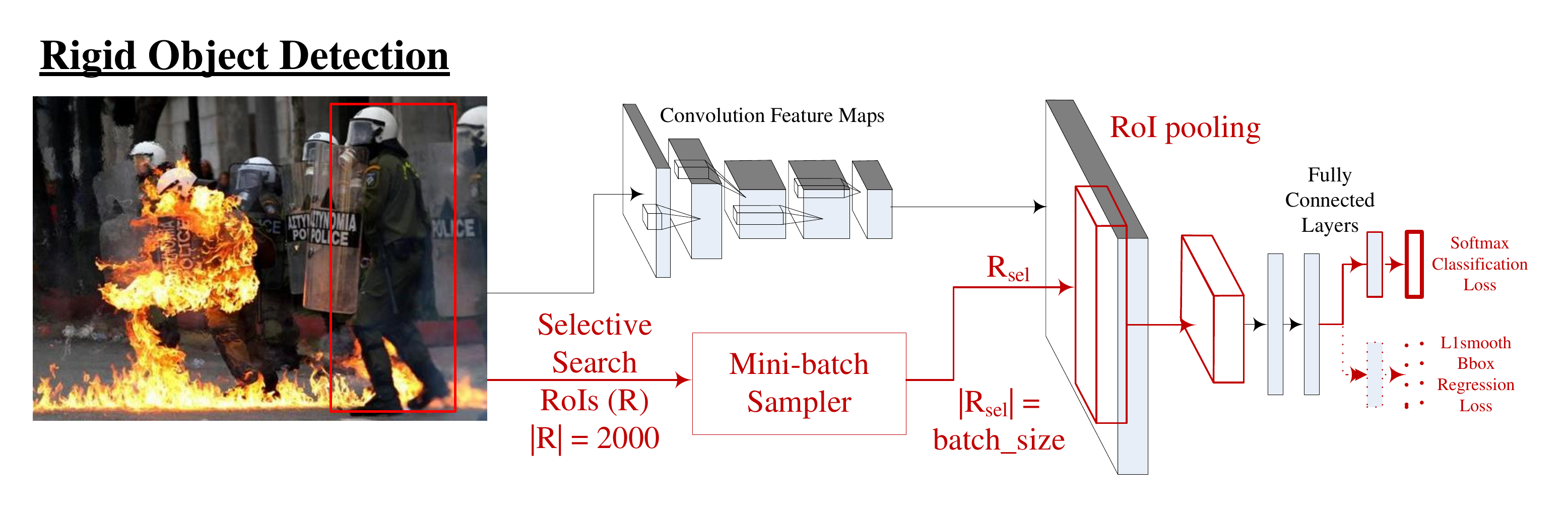}
\subcaption{}
\label{fig:robj_detection}
\end{subfigure}
\begin{subfigure}[htb]{\linewidth}
\includegraphics[trim=7mm 15mm 7mm 5mm,width=\linewidth]{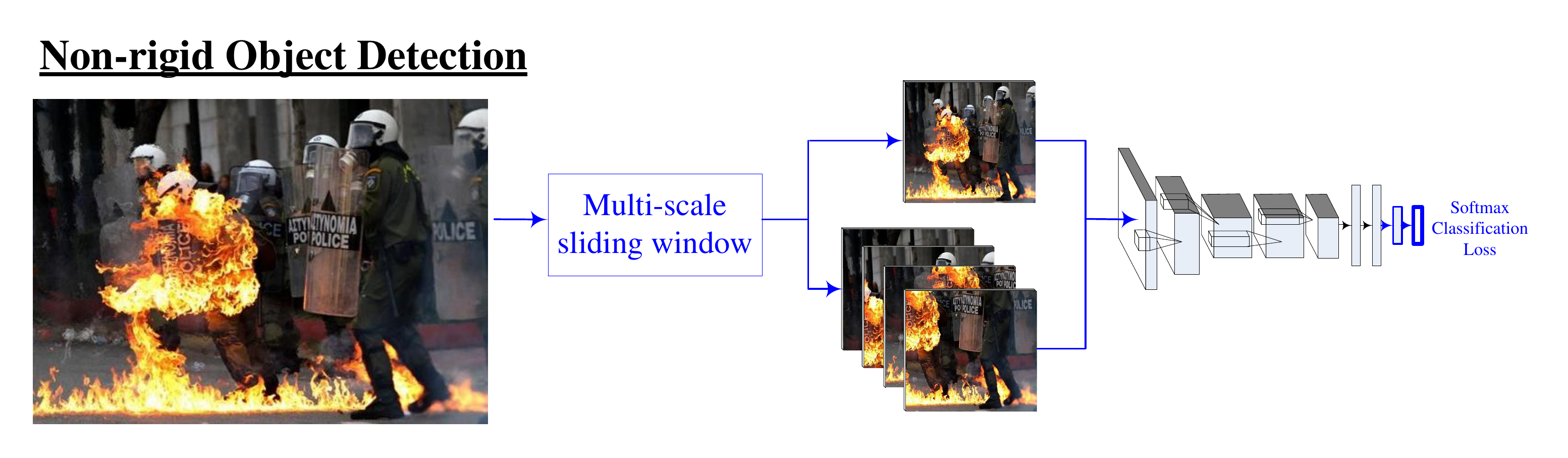}
\subcaption{}
\label{fig:nrobj_detection}
\end{subfigure}
\begin{subfigure}[htb]{\linewidth}
\includegraphics[trim=7mm 20mm 7mm 5mm,width=\linewidth]{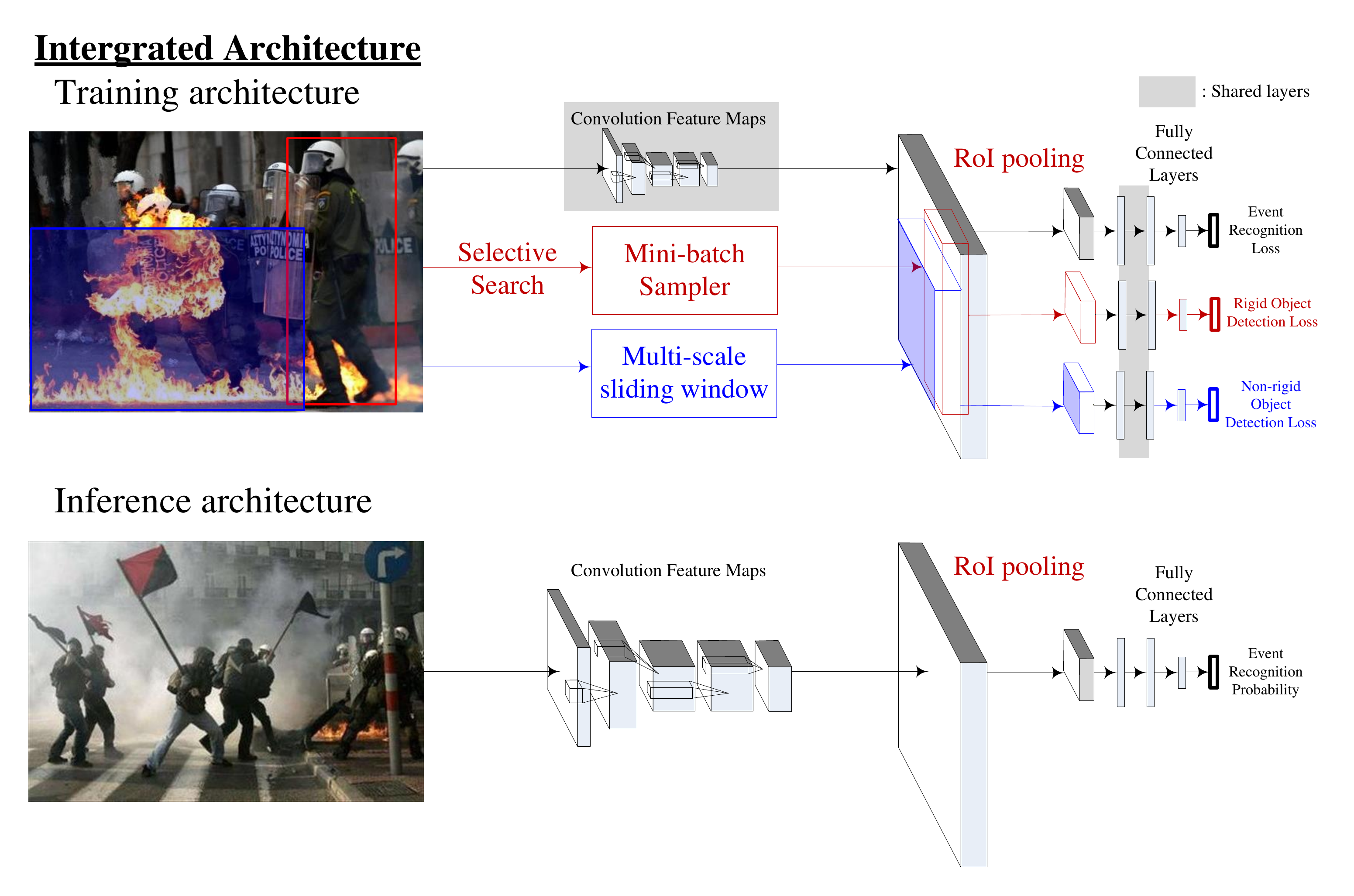}
\subcaption{}
\label{fig:intergrated_CNN}
\end{subfigure}
\vspace{-0.4cm}
\caption{{\bf IOD-CNN architecture.} (top three) Architectures for three separate tasks before the integration (bottom) A novel architecture which integrates event classification with rigid and non-rigid object detection} \label{fig:architecture}
\end{figure}

\noindent{\bf Event Recognition.}
We use a common classification architecture, known as ConvNet, for event recognition. As shown in Fig. \ref{fig:event_recognition}, the network typically consists of a number of convolutional layers followed by a few FC layers. The input is an image with predefined fixed width and height (for both training and testing), while the output is the softmax probability estimates over all of the classes.\\

\noindent{\bf Rigid Object Detection.}
As shown in Fig. \ref{fig:robj_detection}, the Fast R-CNN (FRCN) \cite{RGirshickICCV2015} was chosen to perform the rigid object detection. Unlike the deep ConvNet which requires resized images as input, the original FRCN architecture takes in a full image as input and passes it through a series of convolutional layers to generate a feature map. This map along with the object proposals (approximately 2k) generated by selective search are then fed into a Region of Interest (RoI) pooling layer. The output from the RoI pooling is fed into the FC layers which are followed by two output layers: one for the softmax class-wise probability estimation and the other for the bounding box regression.

The bounding box regression is removed from our architecture (dotted box in Fig. \ref{fig:robj_detection}) because the primary task (event recognition) does not benefit from it. This is because the power of bounding box regression in the original FRCN is exhibited in the post-processing which is separate from the learning process. We have experimentally observed that when object detection is learned along with the bounding box regression in a multi-task scheme, the performance degrades unless the bounding box regression post-processing is existent. In short, incorporating the bounding box regression into our architecture will have a negative effect for the primary task, as there is no chance to perform the post-processing to make up for the loss.\\

\begin{figure}[t]
\includegraphics[trim=8mm 10mm 8mm 0mm, width=\linewidth]{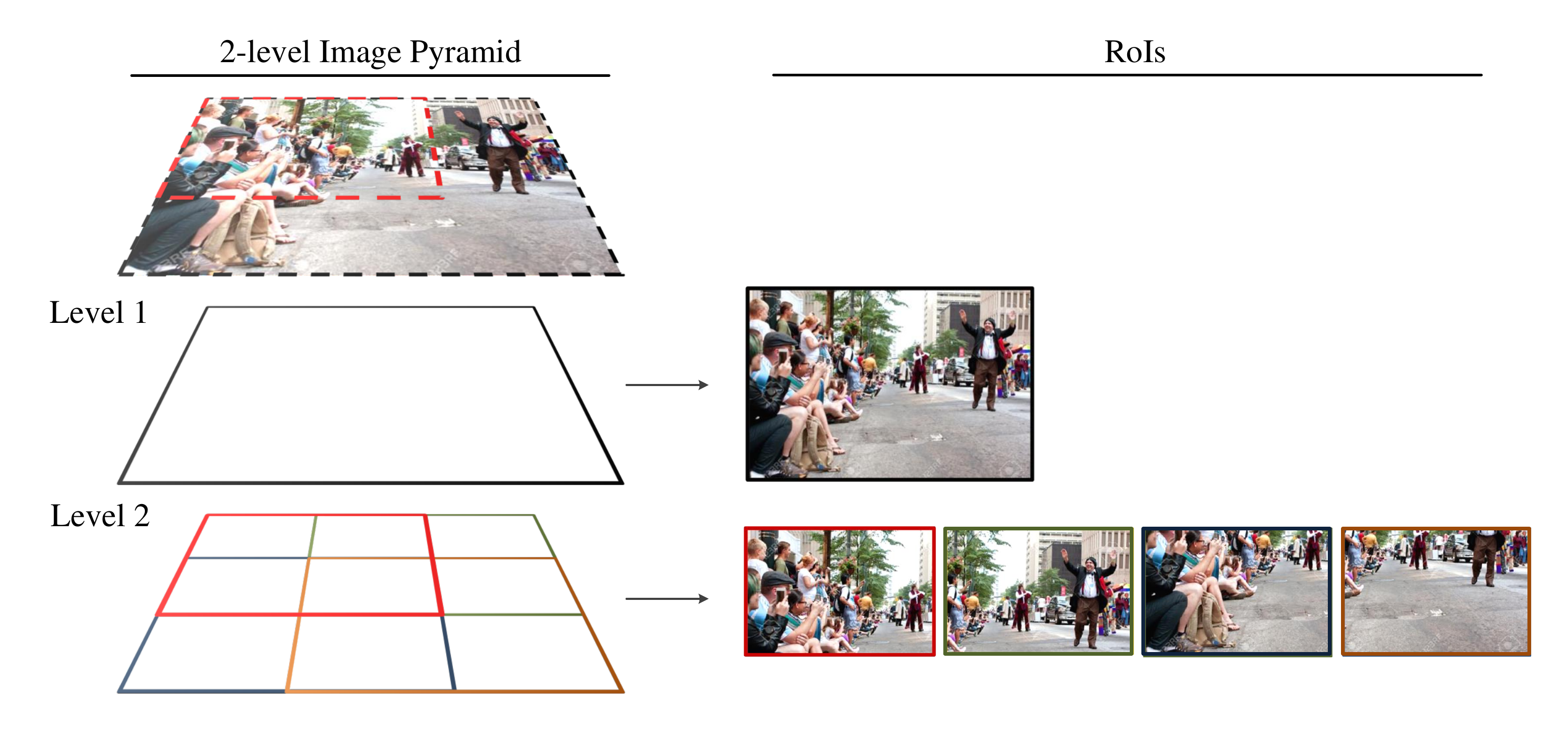}
\caption{Multi-scale sliding window for non-rigid object detection}\label{fig:multiscale}
\end{figure}

\noindent{\bf Non-rigid Object Detection.} Modeling the ``objectness'' for objects with non-standard or non-rigid shape, such as smoke or fire, is not only difficult but also computationally expensive. Thus, instead of using selective search (i.e., used for rigid object detection), which can be considered a fine scanning method, we use a multi-scale sliding window strategy as shown in Fig. \ref{fig:multiscale}. For one input image, five RoIs are generated: one covering the whole image region and the other four covering the four overlapping regions with 2/3 height and 2/3 width of the whole region. These five RoIs are fed into the network as shown in Fig. \ref{fig:nrobj_detection}. \\

\noindent{\bf Integrating the Different Architectures.}
The unified network for training and inferencing are shown in Fig. \ref{fig:intergrated_CNN}. The training architecture consists of a series of convolutional layers, a RoI pooling layer, and three separate modules responsible for event recognition, rigid-object detection and non-rigid object detection, respectively. Each module consist of one shared and two non-shared FC layers. For testing, only the components responsible for the event recognition (primary task) are included in the architecture.

The training network takes an input image and passes it through a series of convolutional layers until it reaches the RoI pooling layer. At the same time, the input image goes through two different sample generators: the selective search and the multi-scale sliding window search, generating samples for rigid and non-rigid object detection, respectively. The output of the convolutional layers along with the outputs of the two sample generators are fed into the shared RoI pooling layer. The three task-specific streams go through the FC layers. Each stream is connected to an appropriate loss function at the end.

The effective integration of these architectures was made possible by sharing the convolutional layers and the first FC layer ({known as \it fc6}) which are learned to serve all three tasks. Note that, the other two ``task-specific'' FC layers ({\it fc7} and {\it fc8}) are learned separately for different tasks. By sharing these layers, we provide each task a means to associate the information from the other tasks. In the experiments (Section \ref{sec:Experiments}), we show that the performance of our primary task is indeed boosted by this integration. In addition, although the RoI pooling layer is not a layer to be learned, it serves a crucial role in allowing full-size input images to be fed into the convolutional layers without resizing.

It is noted in \cite{AkrizhevskyNIPS2012} that the first convolutional layer ({\it conv1}) is more generic and task independent than other convolutional layers. In our case, we share a similar philosophy, but we also show that the network can be better learned when the overall set of convolutional layers is shared and learned together between the semantically-related tasks.

\subsection{Learning the Unified Network}
We have found the network introduced by Krizhevsky et al. \cite{AkrizhevskyNIPS2012} suitable for the single-task event recognition architecture. 
To label the RoIs (for training purpose) in the rigid and non-rigid object detection, we have used 0.5 and 0.2 as the thresholds for the intersection over union (IoU) metric. While the {\it fc6} and {\it fc7} are fine-tuned, the weights for {\it fc8} are initialized by samples from a Gaussian distribution with zero mean and 0.1 standard deviation.

For every iteration, a batch of two images is used. We made sure that each batch is comprised of one sample with a benign label (a normal scene) and one with a malicious one (which would draw attention of law enforcement). For training the rigid object detection, the network takes 64 RoIs from each image which is the selected subset of the initial RoI set provided by the selective search. For event recognition and non-rigid object detection, 1 and 5 RoIs are generated per image, thus 2 and 10 RoIs are used as one batch, respectively.\\

\noindent{\bf Cascaded Optimization.}
One technical challenge in learning the IOD-CNN is selecting the appropriate learning parameters. Naively using the parameters optimized for one of the three modules may not be suitable for acquiring the best performance out of the unified network. For the event recognition and non-rigid object detection, all the RoIs acquired from one image are used for one batch. However, for the rigid object detection, approximately 2000 RoIs are generated per image and only the subset of those RoIs (i.e., 64 for malicious and 64 for benign) are used per batch. To allow more training iterations for the rigid object detection module, we have employed a three step cascaded optimization strategy. The initial CNN network is first trained on the Places Dataset \cite{BZhouNIPS2014}.  Then only the rigid object detection module is learned/fine-tuned on the target dataset (Malicious Crowd Dataset in Section \ref{ssec:Dataset}) using the learning rate of 0.01, 30k iterations, and the step size of 20k. Lastly, the unified network (i.e., IOD-CNN with all the modules) is trained with the learning rate of 0.0001, 12k iterations, and the step size of 8k. 

\section{Experiments}
\label{sec:Experiments}

\begin{figure}[t]
\includegraphics[trim=8mm 45mm 8mm 0mm, width=\linewidth]{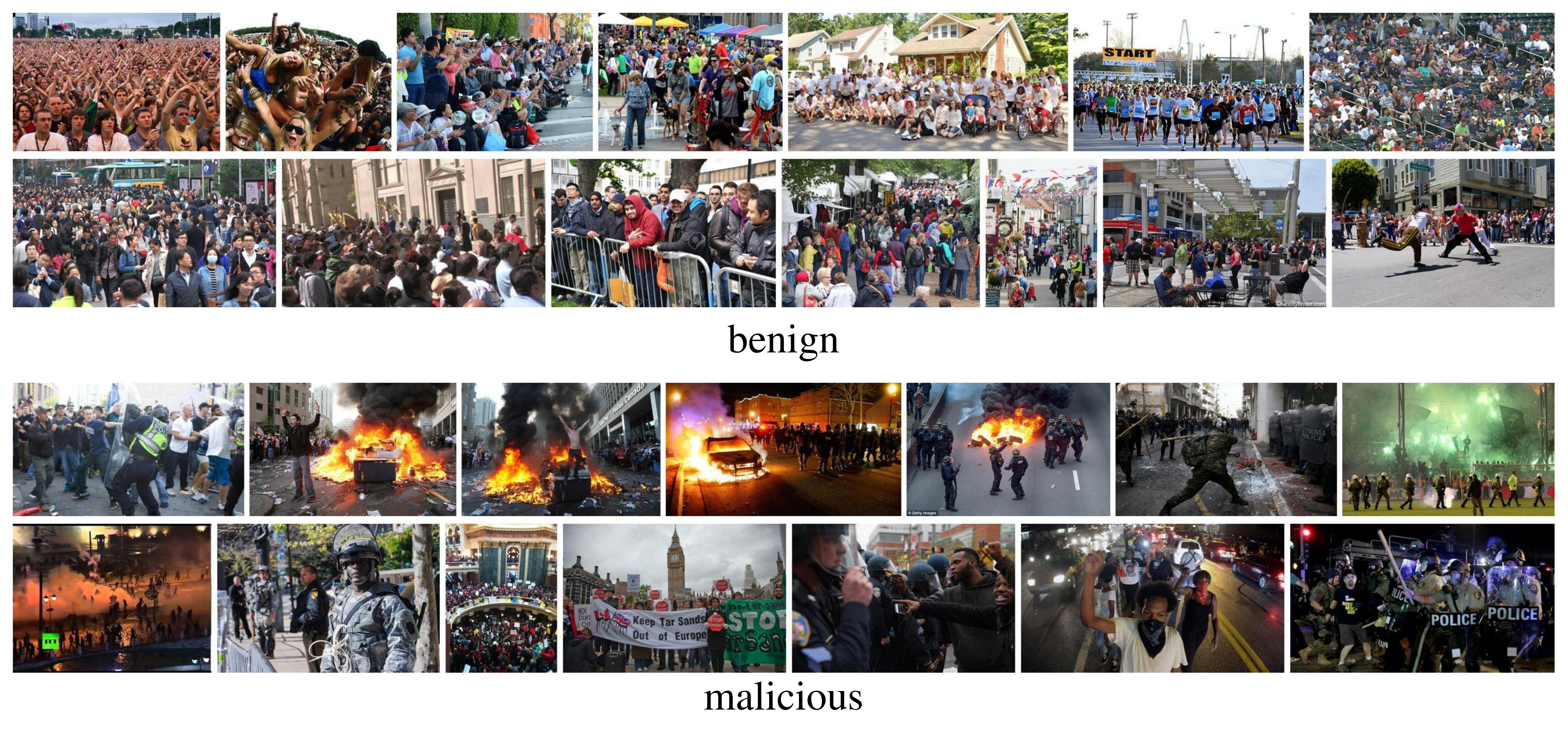}
\caption{Sample images from the Malicious Crowd Dataset with two classes: benign and malicious events} \label{fig:dataset}
\end{figure}

\subsection{Dataset}
\label{ssec:Dataset}
To demonstrate the effectiveness of our architecture, we use the Malicious Crowd Dataset introduced in \cite{JLevisArxiv2016}. This dataset was chosen as it contains not only the crowd event images but also the ground truth labels for relevant objects which are suitable for testing our architecture which requires both image classification and object detection. The dataset contains 1133 crowd images, equally split into {\it malicious} and {\it benign} classes. Sample images are shown in Fig. \ref{fig:dataset}. The {\it malicious} label is said to have been assigned to an image when the scene would be alarming to a passerby or a law enforcement personnel. For both classes, the images contain two different types of objects: rigid (e.g., cars) and non-rigid (e.g., smoke). The dataset also provides the bounding boxes of the frequently appearing ``{\it malicious}-related'' objects which are police, helmet, car, fire, and smoke. The bounding boxes are used to train and evaluate the rigid and non-rigid object detection. Details on how the objects are selected is given in \cite{JLevisArxiv2016}.

\subsection{Performance Evaluation}
\label{ssec:PerformanceEvaluation}

\begin{table}
\caption{{\bf Event recognition} average precision (AP). All methods use \cite{AkrizhevskyNIPS2012} as the baseline architecture. Task: {\bf E}: Event Recognition, {\bf R}: Rigid Object Detection, {\bf N}: Non-rigid Object Detection. \cite{JLevisArxiv2016}* reproduces the result of \cite{JLevisArxiv2016} with our network learning strategy. }
\vspace{-0.6cm}
\setlength{\tabcolsep}{10.5pt}
\begin{center}
\begin{tabular}{llc}
\specialrule{.15em}{.05em}{.05em}
Method & Tasks & AP \\
\specialrule{.15em}{.05em}{.05em}
Single CNN \cite{JLevisArxiv2016} & - & 72.2 \\
Single CNN \cite{JLevisArxiv2016}* & - & 82.5 \\
Single CNN+RoI pooling & - & 90.2 \\\hline
IOD-CNN & E, R & 91.8 \\
IOD-CNN & E, N & 91.9 \\
IOD-CNN & E, R, N & \underline{\bf 93.6} \\
\specialrule{.15em}{.05em}{.05em}
2 CNNs\&DPM+Score Fusion \cite{JLevisArxiv2016} & - & 77.1 \\
OS-CNN+fc7\&TDD Fusion \cite{LWangICCVW2015} & - & 92.9\\
\hline
3 Separate CNNs+Score Fusion & - & 92.9 \\
IOD-CNN+Score Fusion & E, R, N & 93.9\\
IOD-CNN+fc7\&TDD Fusion & E, R, N & \underline{\bf 94.2} \\
\specialrule{.15em}{.05em}{.05em}
\end{tabular}
\end{center}
\label{table:eventRecognition}
\end{table}

\begin{table}[t]
\caption{{\bf Single task versus multitask performance.} {\bf C}: Classification, {\bf D}: Detection, {\bf R} and {\bf N} used mean average precision (mAP) as the evaluation metric. }
\vspace{-0.7cm}
\setlength{\tabcolsep}{5.5pt}
\begin{center}
{\small
\begin{tabular}{lccc}
\specialrule{.15em}{.05em}{.05em}
Method & C/D & Single-task (AP/mAP) & Multi-task (AP/mAP) \\
\specialrule{.15em}{.05em}{.05em}
E & C & 90.2 & \underline{{\bf 93.6}} \\
R & D & \underline{{\bf 11.8}} & 11.0 \\
N & C & 27.7 & \underline{{\bf 82.1}} \\
\specialrule{.15em}{.05em}{.05em}
\end{tabular}
}
\end{center}
\label{table:singlemultitask}
\end{table}

We have carried out a set of experiments to demonstrate how our architecture integration approach can boost event recognition performance to a new state-of-the-art. For all the experiments described in this subsection, we have used the Malicious Crowd Dataset briefly described in the previous subsection.

The first six rows of Table \ref{table:eventRecognition} show that IOD-CNN without any fusion processing outperforms all the baseline single CNNs. The results indicate that integrating rigid (R), non-rigid (N), or both (R,N) object detections into the network all show superior performance, and integrating both works the best. Moreover, we verify that incorporating the RoI pooling layer which allows the input images of arbitrary size, increases the performance.

In the last five rows of Table \ref{table:eventRecognition}, we have also compared IOD-CNN with two baselines  \cite{JLevisArxiv2016,LWangICCVW2015} which use multiple CNNs and exploit fusion strategies. To make a fair comparison with the baselines, we use the same fusion techniques, i.e., score fusion \cite{HLeeWACV2016} and fc7\&TDD fusion. To generate a two stream network, we prepared two networks pretrained on the ImageNet \cite{JDengCVPR2009} and the Places \cite{BZhouNIPS2014} Datasets, as in \cite{LWangICCVW2015}. By applying the same score fusion or fc7\&TDD fusion used in \cite{JLevisArxiv2016} and \cite{LWangICCVW2015}, the performance of {\it pre-fusion} IOD-CNN is improved by 0.3 and 0.6 AP, respectively. This indicates that the early-sharing of  the  network  layers (convolutional and one FC)  is  complementary to the late fusion in terms of the  performance. The IOD-CNN with either of the fusion strategies outperforms all the baselines and the case where 3 separate CNNs (E,R,N) are score-fused.

We have also carried out an experiment to analyze how the performance of each task changes when all the tasks are learned together using the IOD-CNN. Table \ref{table:singlemultitask} shows that the event recognition and the non-rigid object detection performance is boosted when learned together. Notably, the non-rigid object detection performance improved drastically by almost three fold.





\section{Conclusion}
\label{sec:Conclusion}

We presented a novel unified deep CNN architecture which integrates architecturally different, yet semantically-related networks for different tasks to enhance the performance of event recognition. The experimental results show that each of the newly incorporated architecture components are crucial in boosting the performance. The architecture which integrates the two object detections with the event recognition outperforms the previous object-aware event recognition CNNs. As one unified network is learned in an end-to-end fashion, the training can also be performed more efficiently. Moreover, the performance of our architecture can be further improved by appending a late fusion approach. This indicates that the within-network sharing of the layers is complementary to the late fusion.


\bibliographystyle{IEEEbib}
\bibliography{strings,refs}

\end{document}